# Managing Uncertainty in Rule Based Cognitive Modeling

Thomas R. Shultz

Department of Psychology, McGill University

1205 Penfield Avenue, Montréal, Québec, Canada H3A 1B1

E-mail: ints@musicb.mcgill.ca

## ABSTRACT

An experiment replicated and extended recent findings on psychologically realistic ways of modeling propagation of uncertainty in rule based reasoning. Within a single production rule, the antecedent evidence can be summarized by taking the maximum of disjunctively connected antecedents and the minimum of conjunctively connected antecedents. The maximum certainty factor attached to each of the rule's conclusions can be scaled down by multiplication with this summarized antecedent certainty. Heckerman's modified certainty factor technique can be used to combine certainties for common conclusions across production rules.

## INTRODUCTION

Rule based systems have proven to be successful techniques for the computational modeling of human reasoning. They can be used to model human procedural knowledge in a convenient, homogeneous, modular fashion that is consistent with a great deal of psychological evidence (Klahr, Langley, & Neches, 1987).

Curiously, very few of the rule based human simulations employ techniques for representing and propagating uncertainty. Although it is widely acknowledged that much of human knowledge is uncertain, it is in the field of artificial intelligence that research on the representation and management of uncertainty in rule based reasoning has been focused (Kanal & Lemmer, 1986; Hink & Woods, 1987). Most of the work on uncertainty in artificial intelligence has so far been normative, stressing issues of mathematical correctness and effectiveness. The approach taken in this paper is not normative, but descriptive. It seeks to determine how ordinary reasoners propagate uncertainty in the context of rule based reasoning.

The problem of uncertainty in rule based architectures can be broken into two sub-problems. One concerns how the possibly uncertain evidence in a rule's antecedents affects the rule's conclusions. Consider the general case of a production rule with $i$ antecedents and $j$ conclusions.

```
IF  antecedent₁
    antecedent₂
    ⋮
    antecedentᵢ
THEN conclusion₁ (maximum-cf₁)
     ⋮
     conclusionⱼ (maximum-cfⱼ)
```

Attached to each of the rule's $j$ conclusions would typically be a numerically represented maximum certainty factor. If the evidence contained in the rule's antecedents is believed with perfect certainty, then each conclusion$_j$ would be drawn with its maximum-certainty-factor$_j$. However, in the general case, the evidence in each of the rule's antecedents is believed with varying degrees of certainty. How should the uncertainty of antecedent evidence be summarized? And how should this summarized antecedent certainty affect the maximum certainty factor of each conclusion? Slightly complicating the first question is the fact that the antecedents can be connected either conjunctively or disjunctively. With conjunctive connectives, all of the antecedents must hold in order for the rule to fire. For disjunctive connectives, satisfaction of only a single antecedent enables the rule to fire.

The other uncertainty sub-problem in rule based systems concerns the issue of combining evidence across different rules with the same conclusion. Imagine that a particular conclusion exists in more than one rule. As rules fire, their conclusions come to be believed with varying degrees of certainty, as just outlined. How should these certainties be combined in cases where a previously fired rule has an overlapping conclusion with a newly fired rule? This is not a problem in deterministic production systems that do not quantify uncertainty since they typically avoid drawing the same conclusion more than once. However, it is a problem in any production system that attempts to propagate quantitative uncertainties as its rules fire.

Solution of these two sub-problems is critical for rule based efforts to model human cognition because algorithms implementing a solution to each of these two sub-problems are typically invoked every time a rule with quantitative representation of uncertainty fires. If these algorithms lack psychological validity, simulation errors will tend to accumulate and compound as rules fire.

A recent psychological experiment evaluated a number of different solutions to these problems (Shultz, Zelazo, & Engelberg, 1989). In a study of the problem of propagating uncertainty within a rule, subjects learned hypothetical production rules in which each of three antecedents was believed with varying degrees of certainty. A maximum certainty factor for each rule's conclusion was also specified. Subjects were then asked to rate the certainty with which they believed the rule's conclusion. Seven different quantitative models for summarizing the antecedent certainty and two models for using this summarized antecedent certainty to scale down the maximum certainty factor in the rule's conclusion were evaluated.

The results indicated that the best fitting model for summarizing antecedent certainties was the so-called *maximin* model, which took the maximum certainty factor from disjunctively connected antecedents and the minimum certainty factor from conjunctively connected antecedents. The better technique for scaling down the maximum certainty factor in the conclusion was to multiply the maximum certainty factor by this summarized antecedent certainty factor.

Among the alternative, less successful models for summarizing antecedent certainties were minimum, maximum, product, sum minus overlap, mean, and median. The best-fitting maximin model was a hybrid model combining the maximum and minimum models. Another assessed hybrid model combined the product and sum minus overlap models. It used the product of conjunctively connected antecedent certainties and the sum minus overlap of disjunctively connected antecedent certainties.

The less successful model for scale down the maximum certainty factor in the rule's conclusion was a model that computed the mean of this value and the summarized antecedent certainty.

In a second experiment, Shultz et al. (1989) compared five different quantitative models for combining certainties across rules with the same conclusion. Here, subjects learned two rules relevant to the same conclusion, each of which had antecedent evidence of varying certainty. They were asked to rate the certainty of the common conclusion. Two models for scaling down the conclusion's maximum certainty factor by the antecedent certainty were also tested, with the same result as in the first experiment. That is, scaling was better approximated by multiplication than by computing the mean. Two models for combining certainties across rules fit the data equally well: the classic certainty factor approach used in MYCIN (Shortliffe, 1976) and Heckerman's (1986) modified certainty factor approach, which sums the certainties contributed by each of two rules and divides by 1 plus their product. Heckerman's model was favored over the tri-partite classic certainty factor approach because it employs a simpler, more unified formula.

Heckerman (1986) showed that both his modified certainty factor model and the classic certainty factor approach are valid probabilistic interpretations of certainty factors, under the assumptions that the evidence provided by the rules is conditionally independent and that the rule base forms a tree structure. Grosof (1986) showed that Heckerman's modified certainty factor model is equivalent to a special case of Dempster-Shafer theory. We found that the Dempster-Shafer technique (as described in Gordon & Shortliffe, 1984) yielded identical results to the classic certainty factor model when both of two rules either confirmed or disconfirmed a conclusion but differed slightly when one rule was confirming and the other was disconfirming.

Other, less successful models for combining certainties across rules included those that computed the mean, maximum, or minimum of the two certainties.

A criticism addressed to these conclusions at the meeting in which they were first presented was that the problem had been broken into unrealistically small parts. It was argued that the favored algorithms might not work on more natural problems that required full integration of all of the algorithmic elements. The purpose of the present study was to address this critique by using problems that required full integration of the algorithms.

## METHOD

The subjects were 44 university student volunteers. On each of 12 items, they learned two production rules said to express independent sources of evidence for event X. For example,

> If events A, B, and C all happen, then event X is highly certain to happen.
> Event A is highly certain to happen.
> Event B moderately certain to happen.
> Event C is slightly certain to happen.

> If event D, E, or F happens, then event X is moderately certain not to happen.
> Event D is highly certain to happen.
> Event E moderately certain to happen.
> Event F is slightly certain to happen.

In each item, the subject was asked to rate the certainty of event X happening by placing a slash on a 16 cm line with *certain not to happen* anchoring the left end, *certain to happen* anchoring the right end, and *uncertain* labelling the midpoint.

Across the 12 rule items there was systematic variation in the type of connective for the antecedents (conjunctive vs. disjunctive) and the certainty and direction of the conclusion (highly vs. moderately certain to happen vs. not-happen).

As in the original experiments (Shultz et al., 1989), additional items were presented at the end of this questionnaire in order to calibrate each subject's use of the certainty descriptors *highly*, *moderately*, and *slightly* certain. For each of these certainty descriptors, the subject was asked to place a slash on a 16 cm line to represent the described degree of certainty. The calibration line was anchored at one end with *uncertain* and at the other end with *completely certain*.

Two different forms of the questionnaire were created by using two different random orders of the same 12 items. Each of the two forms was given to 22 subjects.

The letters symbolizing events in these items (A, B, C) were not instantiated with more realistic events. It was part of the research strategy to begin with these abstract items before investigating the effects of context on reasoning under uncertainty.

## RESULTS

Responses to the calibration items were converted to certainty factors by measuring the placement of the slash to the nearest 1/2 cm and dividing by 16. These calibrated certainty factor values were then used to generate model predictions for each subject. Responses to the 12 rule items were converted to certainty factors by measuring the placement of the slash to the nearest 1/2 cm, subtracting 8 from this value, and dividing by 8. These conversions insured that all ratings would be on the same certainty factor scale ranging from -1 (completely certain not to happen) to 1 (completely certain to happen), with 0 as the uncertain midpoint.

Two different models were used to predict the human data. One employed the combination of best fitting algorithms from the previous experiments (Shultz et al., 1989): *maximin* summarizing, multiplication scaling, and Heckerman's modified certainty factor technique. For convenience, I call this the MMH model. Because the previous experiments had included so many unsuccessful models, it was thought unnecessary to replicate all of those models here. Just one model, suggested by the spontaneous comments of some subjects in the current experiment was used as a foil to MMH. This was what I call the *mean* model since it summarizes antecedent evidence within a rule by taking the mean certainty factor, scales down the maximum certainty factor in the conclusion by taking the mean of the maximum certainty factor and the summarized antecedent certainty factor, and combines certainties across rules by taking the mean. Mean models were included in our previous experiments, but were not found to be among the best fitting. Nonetheless, taking the mean is not an unreasonable strategy for people faced with a complex task of combining numerical estimates. Moreover, some subjects do insist that that is what they do, even if their data do not always confirm this introspection.

For each subject, both the MMH and the mean models were used to generate the subject's predicted ratings of the certainty of event X happening on the 12 items. Each subject's own calibration ratings of the relevant certainty descriptors were used to generate these predictions. Then the predictions generated by each of the two models were correlated with the subject's actual certainty ratings. Many of these correlation coefficients were statistically significant at $p < .05$ even with only 10 *df*. Thirty-nine of the 44 MMH correlations were significant and 35 of the 44 *mean* correlations were significant.

To evaluate the relative success of the two tested models, the correlation coefficients were subjected to an analysis of variance in which the form of the questionnaire served as a between subject factor and model served as a within subject factor. This analysis yielded only a main effect for model, $F(1, 42) = 25.41, p < .001$. The mean correlation for the MMH model was .75, and that for the mean model was .65. The median correlations for the two models were .83 and .71, respectively.

The mean and median correlations between the predictions generated by the two tested models were .877 and .879, respectively. With regard to mean correlations, the partial correlation between the mean model and actual data, with the MMH model partialled out, was -.03. With regard to median correlations, the partial correlation between the mean model and actual data, with the MMH model partialled out, was -.07.

## CONCLUSIONS AND DISCUSSION

The predictions generated by the MMH model were significantly better than those generated by the mean model. This replicates our previous results and extends them to a more realistic scenario in which all of the relevant algorithms need to function. A realistic way to simulate the data of human subjects in rule based propagation of uncertainty is to use *maximin* summarizing, multiplication scaling, and Heckerman's modified certainty factor model for combining evidence across rules.

All of the relation between the mean model and subject data can be attributed to the fact that both were highly correlated with the predictions of the MMH model. With the MMH model predictions partialled out, the correlation between the mean model and subject data disappeared.

A limitation of this approach is that it is restricted to reasoning with abstract, de-contextualized material. Future research will be necessary to extend these effects to more realistic items. As that happens, theoretical ideas about the impact of context on reasoning under uncertainty can be developed and contextualized results can be compared to the those generated in these abstract settings.

Numerical approaches to reasoning under uncertainty (of which the certainty factor approach is an example) have been criticized for being psychologically unrealistic. Perhaps the most convincing version of this argument is provided by a recent investigation of the diagnostic reasoning of three expert physicians (Kuipers, Moskowitz, & Kassirer, 1988). Think-aloud and cross-examination protocols were obtained from these physicians as they analyzed a difficult medical case possessing numerous uncertainties. The physicians' references to likelihoods were typically categorical or ordinal and only rarely in explicitly numerical form. However, the relative rarity of numerically expressed likelihoods in verbal protocols does not preclude a mathematical account of the propagation of uncertainty in human reasoning. There are successful mathematical accounts of many phenomena in which the mathematics are not consciously represented. The present results, coupled with those of Kuipers et al. (1988), suggest that adequate mathematical models of reasoning under uncertainty can be achieved even if the combination formulas are not part of the subject's explicit knowledge.

Although quite successful in accounting for human data, the modified certainty factor approach advocated here does suffer from a subtle flaw that needs correcting. The original certainty factor approach for combining disconfirming evidence across rules involves a simple summing of the two certainty factors (Shortliffe, 1976). Later, it was recognized that this could create anomalies when a conclusion with plenty of accumulated support (many rules) could be completely overwhelmed by a single piece of new evidence (one rule). Consequently, the impact of new disconfirming evidence was reduced by dividing the sum by 1 minus the minimum of the two absolute certainty factors. One difficulty is that this is a shot gun solution that minimizes the impact of all new evidence equally, no matter how much evidence has accumulated in favor of the alternate conclusion. A more precise solution would take account of how much evidence had accumulated on each side.

Another difficulty is that the amount of evidence underlying a certainty factor ought to affect combinations of confirming as well as disconfirming evidence. A more general solution would incorporate adjustment for amount of support into all updates, whether confirming or disconfirming. These adjustments might apply, for example, to Heckerman's (1986) modified certainty factor model. Such considerations would not have affected the present results, where there was no accumulation of evidence, but could be quite important in cases where evidence does accumulate. Development of a more precise, yet more generally applicable, solution to these problems is another topic for future investigation.

Learning is another problem for approaches (like the present one) that build relative certainties directly into the knowledge representation. A variety of techniques have been developed to enable production systems to modify their own rule base, essentially by learning new rules (Klahr et al., 1987; Laird, Newell, & Rosenbloom, 1987). It is likely that the learning of maximum certainty factors for rule conclusions would significantly complicate these learning algorithms. Yet the effort required to dynamically adjust certainty factors in rule conclusions as a function of corrective feedback might well be worthwhile. A system that was sensitive to its own relative certainties would be more intelligent and effective than one which ignored such differences.

The certainty factor approach to reasoning under uncertainty can be criticized for not making use of a full, optimal Bayesian analysis, which would take account of prior probabilities of the conclusion and conditional probabilities of the evidence given the conclusion. However, the certainty factor approach can be regarded as Bayesian under restricted assumptions (Heckerman, 1986; Wise & Henrion, 1986). The present results show that the consequences of these assumptions are not disastrous for psychological modeling. Also, psychological experiments have revealed a tendency for people to ignore prior probabilities even if they are well aware of them (Tversky & Kahneman, 1980). Furthermore, it is unlikely that people would typically possess the knowledge of prior and conditional probabilities required by a full Bayesian analysis. The most that ordinary reasoners can realistically be expected to know, at least implicitly, is how certain they are of various instantiated antecedent conditions and the maximum certainty factors contained in the conclusions of their production rules.

Any of the various normative models of reasoning under uncertainty could be candidates for descriptive models of how ordinary people reason. Some of these candidates could conceivably be ruled out as psychological models in terms of the amount of prior knowledge or working memory capacity that they require. Those remaining normative models could serve as inspiration for descriptive psychological investigation. Conversely, it is also possible that psychological evidence might inspire or at least constrain the normative study of reasoning under uncertainty.

## ACKNOWLEDGMENTS

This research was supported in part by a grant from the Natural Sciences and Engineering Research Council of Canada and by the McGill-IBM Cooperative Project.